\documentclass{article}
\usepackage{iclr2015,mathptmx}
\usepackage{graphicx}
\usepackage{amsmath,amsthm,amsfonts, amssymb, dsfont, subfigure, titlesec, authblk}
\newtheorem{theorem}{Theorem}[section]
\newtheorem{lemma}[theorem]{Lemma}

\newcommand{\real}{\mathbb{R}}
\newcommand{\plane}{\mathbb{R}^2}
\newcommand{\glr}{GL_2(\mathbb{R})}
\newcommand{\czxr}{C_0(X,\mathbb{R})}

\newcommand{\oxr}{X \times \mathbb{R}}
\newcommand{\diffczxr}{\mathrm{Diff}(\czxr)}
\newcommand{\homeo}{\mathrm{Homeo}}

\newcommand{\ie}{\emph{i.e.}}
\newcommand{\eg}{\emph{e.g.}}
\newcommand{\id}{\mathds{1}}
\newcommand{\pone}{\textbf{P1}}
\newcommand{\ptwo}{\textbf{P2}}
\newcommand*{\Scale}[2][4]{\scalebox{#1}{$#2$}}%

\titlespacing{\subsubsection}{0pt}{\parskip}{-\parskip}
\titlespacing{\subsection}{0pt}{\parskip}{-\parskip}

\usepackage[final,inline,nomargin,index]{fixme}
\fxsetup{theme=colorsig,mode=multiuser,inlineface=\itshape,envface=\itshape}
\FXRegisterAuthor{sv}{asv}{Suresh}
\FXRegisterAuthor{ap}{aap}{Arnab} 
\iclrconference

\begin{document}

\title{Why does Unsupervised Deep Learning work? - A perspective from Group Theory}
\author[1]{ Arnab Paul \thanks{arnab.paul@intel.com}}
\author[2]{ Suresh Venkatasubramanian \thanks{suresh@cs.utah.edu}}
\affil[1]{Intel Labs, Hillsboro, OR 97124}
\affil[2]{School of Computing, University of Utah, Salt Lake City, UT 84112.}

\renewcommand\Authands{ and }

%
%
%
%
%
%

\maketitle

\vspace{-0.3in}
\begin{abstract}
Why does Deep Learning work? What representations does it capture? How do higher-order
representations emerge? We study these questions from the perspective of group theory, 
thereby opening a new approach towards a theory of Deep learning.

One factor behind the recent resurgence of the subject
is a key algorithmic step called {\em pretraining}: first search for a good generative model for the
input samples, and repeat the process one layer at a time. 
We show deeper implications of this simple principle,
by establishing a connection with the interplay of orbits and stabilizers of group actions. 
Although the neural networks themselves may not
form groups, we show the existence of {\em shadow} groups whose elements
serve as close approximations. 

Over the shadow groups, the pretraining step,
originally introduced as a mechanism to better initialize a network, becomes equivalent to a
search for features with minimal orbits. Intuitively, these features are in a way the {\em simplest}.
Which explains why a deep learning network learns simple features first.
Next, we show how the same principle,  when repeated in the deeper layers, 
can capture higher order representations, and why representation complexity increases
as the layers get deeper.

\end{abstract}

\section{Introduction}
The modern incarnation of neural networks,  now popularly known as Deep Learning (DL),
accomplished record-breaking success in processing diverse kinds of signals -
vision, audio, and text. 
In parallel, strong interest has ensued towards constructing a \emph{theory} of DL. 
This paper opens up a group theory based approach, towards a theoretical 
understanding of DL.

We focus on two key principles that (amongst others) influenced the modern DL resurgence.

\begin{itemize}
\item[(\pone)]  
Geoff Hinton summed this up as follows. 
``In order to do computer vision, first learn how to do computer
graphics". \cite{hinton2007recognize}. In other words, if a network learns a good generative model of its
training set, then it could use the same model for classification.

\item[(\ptwo)]  
Instead of learning an entire network all at once,  learn it one layer at a
time \sverror{Is there a citation for this principle?}.
\end{itemize}

\begin{figure}
\subfigure[General auto-encoder schematic]{\label{fig:aepic1}\includegraphics[width=1.8in, height=1in]{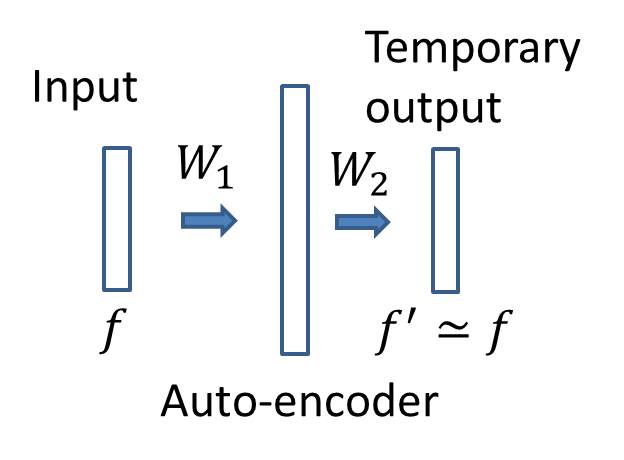}}
\hspace{1in}\subfigure[post-learning behavior of an auto-encoder]{\label{fig:aepic2}\includegraphics[width=1.8in, height=1in]{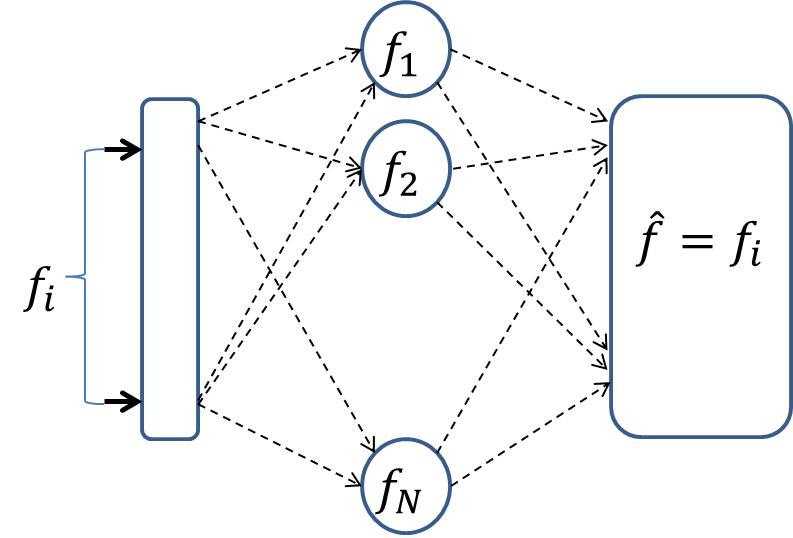}}
\caption{(a) $W_1$ is preserved, $W_2$ discarded \;\;\; (b) Post-learning, each feature is stabilized}
\end{figure}

In each round, the training layer is connected to a temporary output layer and
trained to learn the weights needed to reproduce its input (i.e to solve \pone).
This step -- executed layer-wise, starting with the first hidden layer and
sequentially moving deeper -- is often referred to as pre-training (see
\cite{Hinton06,hinton2007recognize, salakhutdinov2009deep, bengio--dlbook}) and
the resulting layer is called an \emph{autoencoder} \svnote{Is this correct?}.
Figure \ref{fig:aepic1} shows a schematic autoencoder. Its weight set $W_1$ is
learnt by the network. Subsequently when presented with an input $f$, the
network will produce an output $f' \approx f$. At this point the output units
as well as the weight set $W_2$ are discarded.

There is an alternate characterization of \pone.
An autoencoder unit, such as the above, maps an input space to itself.
Moreover, after learning, it is by definition, a \emph{stabilizer}\footnote{ A transformation $T$ is called a stabilizer of an input $f$, if $f' = T(f) = f$.} of the input $f$. 
Now, input signals are often decomposable  into features,
and an autoencoder attempts to find a succinct set of features that all inputs can be decomposed into.
Satisfying \pone  means that the learned configurations can reproduce these features.  
Figure \ref{fig:aepic2} illustrates this post-training behavior. If the hidden units learned features $f_1, f_2, \ldots$, and one of then, say $f_i$, comes back as input, the output must be $f_i$. 
In other words \emph{learning a feature is equivalent to searching for a transformation that stabilizes it}. 

The idea of stabilizers invites an analogy reminiscent of the orbit-stabilizer
relationship studied in the theory of group actions. Suppose $G$ is a group that
acts on a set $X$ by moving its points around (e.g groups of $2 \times 2$
invertible matrices acting over the Euclidean plane). Consider $x \in X$, and
let $O_x$ be the set of all points reachable from $x$ via the group
action. $O_x$ is called an orbit\footnote{Mathematically, the orbit $O_x$ of
  an element $x \in X$ under the action of a group $G$, is defined as the set
  $O_x = \{g(x) \in X | g \in G\}$.}.  A subset of the  group elements may leave $x$
unchanged. This subset $S_x$ (which is also a subgroup), is the stabilizer of
$x$.  If it is possible to define a notion of volume for a group, then there is an inverse relationship between the volumes of
$S_x$ and $O_x$, which holds even if $x$ is actually a subset (as opposed to
being a point). For example, for finite groups, the product of $|O_x|$ and
$|S_x|$ is the order of the group.

\begin{figure}[htbp]
\subfigure[Alternate Decomposition of a Signal]{\label{fig:decompose}\includegraphics[width=3in,height=1.1in]{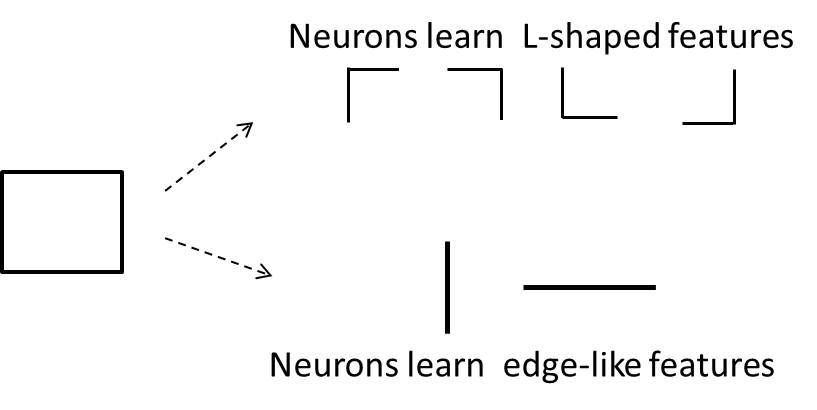}}
\hspace{1cm}
\subfigure[Possible ways of feature stabilization]{\label{fig:sgd}\includegraphics[width=2in, height=1.1in]{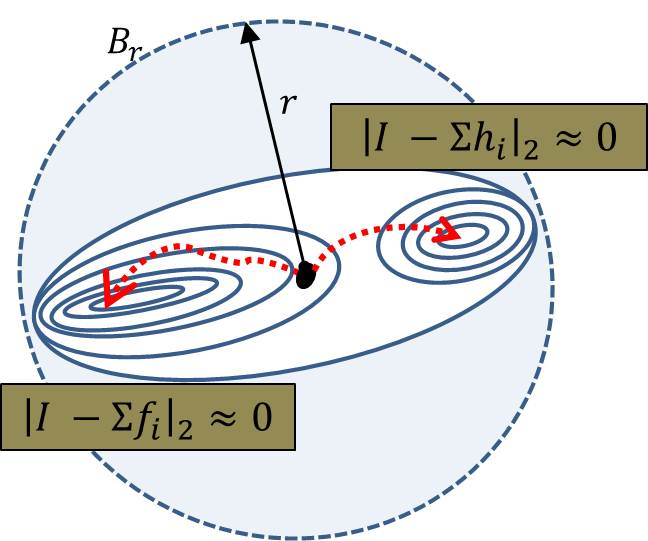}}
\caption{(a) Alternate ways of  decomposing a signal into simpler features. The neurons could potentially 
learn features in the top row, or the bottom row. Almost surely, the {\em simpler} ones (bottom row) are learned.
(b) Gradient-descent on error landscape. Two alternate classes of {\em features} (denoted by $f_i$ and $h_i$) can reconstruct the input $I$
- reconstructed signal denoted by $\Sigma f_i$ and $\Sigma h_i$ for simplicity. Note that the error function
is unbiased between these two classes, and the learning will select whichever set is encountered earlier.}
\end{figure}
The \emph{inverse} relationship between the volumes of orbits and stabilizers
takes on a central role as we connect this back to DL.  
There are many possible ways to decompose signals into smaller features. 
Figure \ref{fig:decompose} illustrates this point: a  rectangle can be
decomposed into L-shaped features or straight-line edges. 

All experiments to date suggest that a neural network is likely to learn the
edges. But why? To answer this, imagine that the space of the autoencoders
(viewed as transformations of the input) form a group.  A batch of learning iterations stops \emph{whenever} a stabilizer is found. 
Roughly speaking, if the search is a
Markov chain (or a guided chain such as MCMC), then 
the bigger a stabilizer, the earlier it will be hit.
The group structure implies that this big stabilizer corresponds to a small orbit. 
Now intuition suggests that the simpler a feature, the smaller is its orbit. 
For example, a line-segment generates many fewer possible shapes\footnote{In fact, one only gets line segments back} under linear 
deformations than a flower-like shape. An autoencoder then should
learn these \emph{simpler} features first, which falls in line with most experiments (see \cite{lee2009convolutional}).

The intuition naturally extends to a many-layer scenario.
Each  hidden layer finding a feature with a big stabilizer.
But beyond the first level, the inputs no longer inhabit the same space as the
training samples. A ``simple" feature over this new space actually corresponds
to a more complex shape in the space of input samples. This process repeats as
the number of layers increases. In effect, each layer learns ``edge-like
features'' with respect to the previous layer, and from these locally simple
representations we obtain the learned higher-order representation.

\subsection{Our Contributions}
\label{sec:our-contributions}
Our main contribution in this work is a formal substantiation of the above
intuition connecting autoencoders and
stabilizers.
First we build a case for the idea that a random search process will find large
stabilizers (section \ref{sec:vol-arg}), and construct evidential examples
(section \ref{sec:gabor}).

Neural networks incorporate highly nonlinear transformations and so our analogy
to group transformations does not directly map to actual networks. However, it
turns out that we can define (Section ~\ref{sec:formulation}) and construct
(Section \ref{sec:group}) \emph{shadow} groups that approximate the actual
transformation in the network, and reason about these instead. 

Finally, we examine what happens when we compose layers in a multilayer
network. Our analysis highlights the critical role of the sigmoid and show how it enables
the emergence of higher-order representations 
within the framework of this theory (Section ~\ref{sec:moduli})

\section{Random walk and Stabilizer Volumes}
\label{sec:vol-arg}

\subsection{Random Walks over the Parameter Space and Stabilizer Volumes}
\label{sec:random-walk}
The learning process resembles a random walk,
or more accurately, a Markov-Chain-Monte-Carlo type sampling. This is already 
known, \eg see (\cite{salakhutdinov2009deep, bengio--dlbook}). 
A newly arriving training sample has no prior correlation with the current state.  
The order of computing the  partial derivatives is also randomized. 
Effectively then, the subsequent minimization step takes off in
an almost random direction , guided by the gradient,
towards a minimal point that {\em stabilizes} the signal. 
Figure \ref{fig:sgd} shows this schematically.
Consider the current network configuration, and its neighbourhood $B_r$ of radius $r$.
Let the input signal be $I$, and suppose that there are two possible
decompositions into features: $f = \{f_1, f_2 \ldots\}$ and $h = \{h_1, h_2
\ldots\}$. 
We denote the reconstructed signal by $\Sigma_i f_i$ (and in the other case, $\Sigma_j h_j$).
Note that these features are also signals (just like the input signal, only simpler).
The reconstruction error is usually given by an error term, such as
the $l_2$ distance ($\| I - \Sigma f_i\|_2$).
If the collection $f$ really enables a good reconstruction of the input - \ie $\| I - \Sigma f_i\|_2 \approx 0$ -
then it is a stabilizer of the input {\em by definition}. 
If there are competing feature-sets, gradient descent will eventually move the configuration to one 
of these stabilizers.

Let $P_f$ be the probability that the network discovers stabilizers for the signals
$f_i$ (and similar definition for $P_h$), in a neighbourhood $B_r$ of radius $r$. 
$S_{f_i}$ would denote the stabilizer set of a signal $f_i$.
Let $\mu$ be a volume measure over the space of transformation. 
Then one can roughly say that 
\[{\frac{ P_f }{ P_h } \propto \frac{\underset{i}\prod{\mu(B_r \cap S_{f_i})}}{\underset{j}\prod{\mu(B_r \cap S_{h_j})}}}\]

Clearly then, the most likely chosen features are the ones with the bigger stabilizer volumes.

\subsection{Exposing the structure of features - From Stabilizers to Orbits}

If our parameter space was actually a {\em finite} group, we could use
the following theorem.

\textbf{Orbit-Stabilizer Theorem}  Let $G$ be a group acting on a set $X$, and $S_f$ be the stabilizer subgroup of an element
$f \in X$. Denote the corresponding orbit of $f$ by $O_f$. Then $|O_f|.|S_f| = |G|$. 

%

For finite groups, the inverse relationship of their volumes (cardinality) is direct; but it does not extend verbatim for continuous groups. Nevertheless, the following {\em similar} result holds:

\begin{equation}
\label{eqn-dims}
	\Scale[0.9]{\dim(G) - \dim(S_f) = \dim(O_f)}
\end{equation}

The dimension takes the role of the cardinality. In fact, under a suitable measure (\eg the Haar measure), a
stabilizer of higher dimension has a larger volume, and therefore, an orbit of smaller volume.
Assuming group actions - to be substantiated later - 
this explains the emergence of simple signal blocks as the learned features in the first layer. We provide some evidential examples by analytically computing their dimensions.




\subsection{Simple Illustrative Examples and Emergence of Gabor-like filters}
\label{sec:gabor}
\begin{figure}
\centering
\includegraphics[width=5in,height=3in]{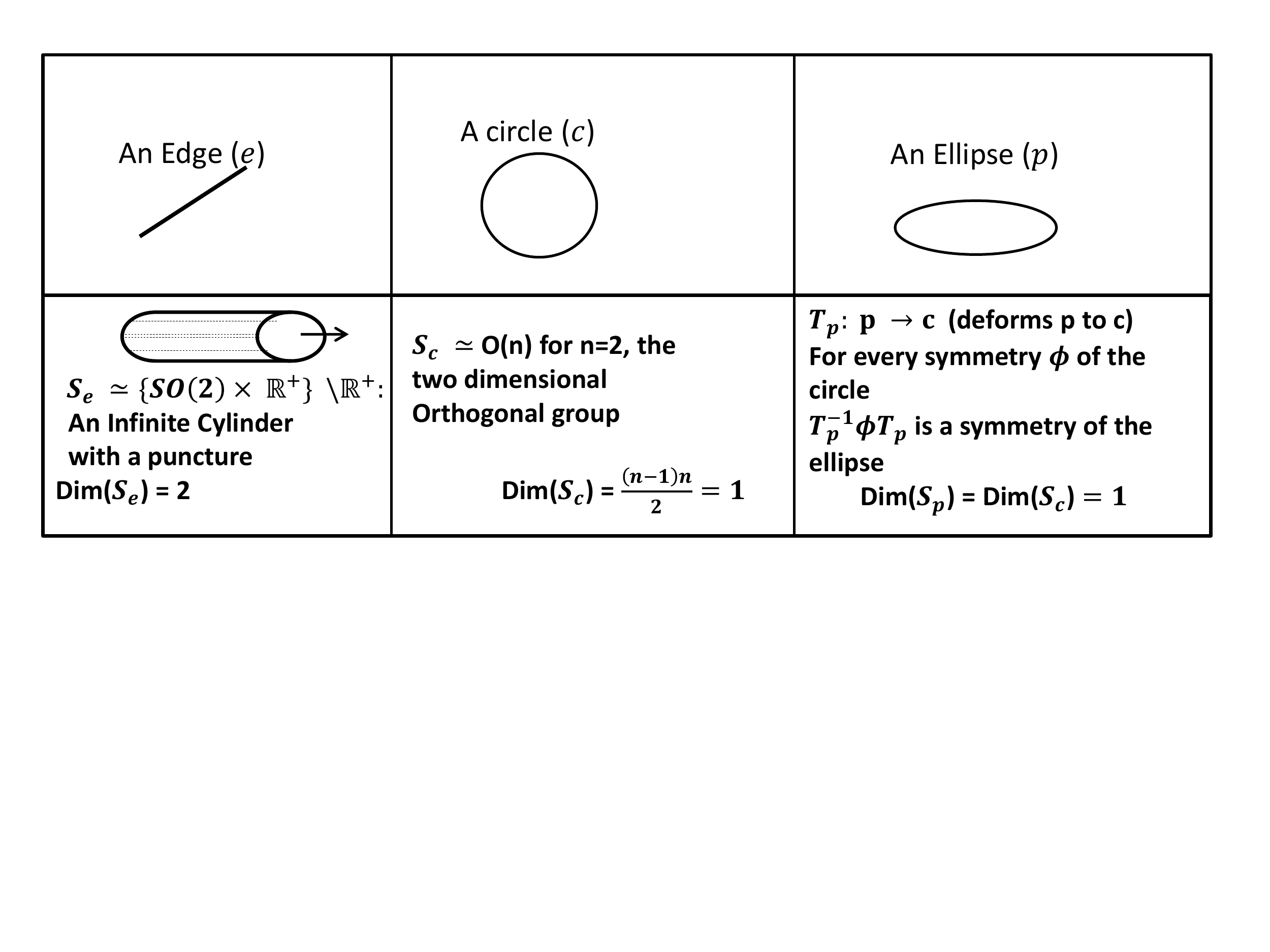}
\vspace{-1.2in}
\caption{Stabilizer subgroups of $GL_2(\mathbb(R))$.  The stabilizer subgroup is of dimension 2, as it is isomorphic
to an infinite cylinder sans the real line. The circle and ellipse on the other hand have stabilizer subgroups
that are one dimensional.}
\label{fig:simp_stab}
\end{figure}

Consider the action of the group $\glr$, the set of {\em invertible} 2D linear transforms, on various 2D shapes. 
Figure \ref{fig:simp_stab} illustrates three example cases by estimating the stabilizer sizes.
\begin{description}
\item[\textbf{An edge(e)}:] For an edge $e$ (passing through the origin), its
  stabilizer (in $\glr$) must {\em fix} the direction of the edge, \ie it must
  have an eigenvector in that direction with an eigenvalue 1.  The second
  eigenvector can be in any other direction, giving rise to a set isomorphic to
  $SO(2)$ \footnote{$SO(2)$ is the subgroup of all 2 dimensional rotations,
    which is isomorphic to the unit circle}, sans the direction of the first
  eigenvector, which in turn is isomorphic to the unit circle punctured at one point.
  Note that isomorphism here refers to topological isomorphism between sets.
  The second eigenvalue can be anything, but considering that the entire circle
  already accounts for every pair $(\lambda, -\lambda)$, the effective set is
  isomorphic to the positive half of the real-axis only.  In summary, this
  stabilizer subgroup is: $S_e \simeq SO(2) \times \real^+ \backslash \real^+ $.
  This space looks like a cylinder extended infinitely to one direction (Figure
  \ref{fig:simp_stab}).  More importantly $\dim(S_e) = 2$, and it is actually a
  non-compact set.

  The dimension of the corresponding orbit, $\dim(O_e) = 2$, as revealed by
  Equation \ref{eqn-dims}.

\item[\textbf{A circle}:] A circle is stabilized by all rigid rotations in the
  plane, as well as the reflections about all possible lines through the
  centre. Together, they form the orthogonal group ($O(2)$) over $\real^2$.
  From the theory of Lie groups it is known that the $\dim(S_c) = 1$.

\item[\textbf{An ellipse}:] The stabilizer of the ellipse is isomorphic to that of a
  circle.  An ellipse can be deformed into a circle, then be transformed by any
  $t \in S_c$ , and then transformed back. By this isomorphism $\dim(S_p) = 1$.

\end{description}
In summary, for a random walk inside $\glr$, the
likelihood of hitting an edge-stabilizer is very high, compared to
shapes such as a circle or ellipse, which are not only compact, but also have one dimension less.
The first layer of a deep learning network, when trying to learn images, almost always discovers
Gabor-filter like shapes. Essentially these are edges of different orientation inside those images. 
With the stabilizer view in the background, perhaps it is not all that surprising after all.


\section{From Deep learning to group action - The mathematical formulation}
\label{sec:formulation}

\subsection{In Search of dimension reduction; The Intrinsic space }
Reasoning over symmetry groups is convenient. Now we shall show that it is possible to 
continue this reasoning over a deep learning network, even if it employs non-linearity.
But first, we discuss the notion of an intrinsic space.
Consider a $N \times N$ binary image; it's typically represented as a
vector in $R^{N^2}$, or more simply in $\{0,1\}^{N^2}$, yet, it is
{\em intrinsically} a two dimensional object.
Its resolution determines $N$, which may change, but that's not intrinsic 
to the image itself. Similarly, a gray-scale image has three intrinsic dimensions - the 
first two accounts for the euclidean plane, and the third for its gray-scales.
Other signals have similar intrinsic spaces. 

We start with a few definitions.

\textbf{Input Space (X)}: It is the original space that the signal inhabits. Most signals of interest are  compactly supported bounded real functions over a vector space $X$. The function space is denoted by  $\czxr = \{\phi | \phi : X \rightarrow \real \}$.

We define \textbf{Intrinsic space} as: $ S = X \times \real$. 
Every $\phi \in \czxr$ is a subset of $S$.  
A neural network maps a point in $\czxr$ to another point in $\czxr$; 
Inside $S$, this induces a deformation between subsets.

An example. A binary image, which is a function $\phi: \plane \rightarrow \{0, 1\}$ naturally corresponds to a subset 
$f_\phi = \{x \in \plane \; \text{such that } \phi(x) =1\}$. Therefore,
the intrinsic space is the plane itself. This was implicit in section \ref{sec:gabor}.
Similarly, for a monochrome gray-scale image, 
the intrinsic space is $S = \real^2 \times \real = \real^3$. In both cases, the input space $X = \real^2$.


\textbf{\em Figure} A subset of the intrinsic space is called a figure, i.e., $f \subseteq S$.
Note that a point $\phi \in \czxr$ is actually a figure over $S$.

\textbf{\em Moduli space of Figures}  One can imagine a space that parametrizes various figures over $S$. We denote this 
by $F(S)$ and call the moduli space of figures. Each point in $F(S)$ corresponds to a figure over $S$.
A group $G$ that acts on $S$, consistently extends over $F(S)$, i.e., for $g \in G, \;\; f \in S$, we get another figure $ g(f) = f' \in F(S)$. 
 
\textbf{\em Symmetry-group of the intrinsic space} 
For an intrinsic space $S$, it is the collection of all invertible mapping $S \rightarrow S$. 
In the event $S$ is finite, this is the permutation group. 
When $S$ is a vector space (such as $\real^2$ or $\real^3$), it is the set $GL(S)$, of all linear invertible transformations.

\textbf{\em The Sigmoid function} will refer to any standard sigmoid function, and be denoted as $\sigma()$.

\subsection{The Convolution View of a Neuron}
\label{sec:convolution}

We start with the conventional view of a neuron's operation.
Let $r_x$ be the vector representation of an input $x$. 
For a given set of weights $w$, a neuron performs the following function (we ommit the bias term 
here for simplicity) - $Z_w ( r_x) =  \sigma(<w,r_x>)$

Equivalently, the neuron performs a convolution of the input signal $I(X) \in \czxr$.
First, the weights transform the input signal to a coefficient in a {\em Fourier-like} space.
\begin{equation}
\label{eqn:tau}
   \tau_w( I) = \int \limits_{\theta \in X} w(\theta)I(\theta)d\theta
\end{equation}
  And then, the sigmoid function thresholds the coefficient
  \begin{equation}
  \label{eqn:zeta}
    \zeta_w( I) = \sigma(\tau_w(I))
\end{equation}

A deconvolution then brings the signal back to the original domain. 
Let the outgoing set of weights are
defined by $S(w,x)$.  The two arguments, $w$ and $x$, indicate that
its domain is the {\em frequency space} indexed by $w$, and range is a set of coefficients
in the space indexed by $x$. For the dummy output layer of an auto-encoder, this space is essentially
identical to the input layer.
The deconvolution then looks like: $\hat{I}(x) = \int_{w } S(w,x)\zeta_w(I)dw$.
%

In short, a signal $I(X)$ is transformed into another signal $\hat{I}(X)$.
Let's denote this composite map $ I \rightarrow \hat{I}$ by the symbol $\psi$,
and the set of such composite maps by $\Omega$, i.e., $\Omega = \{ \psi | \psi : \czxr \rightarrow \czxr\}$.


We already observed that a point in $\czxr$ is a figure in the 
intrinsic space $ S = X \times \real $.
Hence any map $\psi \in \Omega$ naturally induces the following map
from the space $F(S)$ on to itself: $\psi(f) = f^{\prime} \subseteq S$.

Let $\Gamma$ be the space of all deformations of this intrinsic space $S$, i.e., $\Gamma = \{ \gamma | \gamma : S \rightarrow S \}$.
Although $\psi$ deforms a figure $f \subseteq S $ into another figure $f^{\prime} \subseteq S$,
this action does not necessarily extend uniformly over the entire set $S$.  
By definition, $\psi$ is a map $\czxr \rightarrow \czxr$  and not  $\oxr \rightarrow \oxr$.
One trouble in realizing $\psi$ as a consistent $S \rightarrow S$ map is as follows. 
Let $f, g \subseteq S$ so that $ h = f \cap g \neq \emptyset$.
The restriction of $\psi$ to $h$ needs to be consistent
both ways; i.e., the restriction maps $\psi(f)|_h$ and $\psi(g)|_h$ should agree over $h$.  But that's not guaranteed
for randomly selected $\psi, f$ and $g$.  
%

If we can naturally extend the map to all of $S$, then we can translate the questions asked over $\Omega$ to questions over $\Gamma$. 
The intrinsic space being of low dimension, we can hope for easier analyses. In particular,
we can examine the stabilizer subgroups over $\Gamma$ that are more tractable. 



So, we now examine if a map between figures of $S$ can be effectively
captured by group actions over $S$. It suffices to consider 
the action of $\psi$, one input at a time.
This eliminates the conflicts arising from different inputs. Yet, $\psi(f)$ - \ie the action
of $\psi$ over a specific $f \in \czxr$ - is still incomplete with respect to being
an automorphism of $S = \oxr$ (being only defined over $f$). 
Can we then extend this action 
to the  entire set $S$ consistently? It turns out - yes.

\begin{theorem}
\label{thm-automorph}
Let $\psi$ be a neural network, and $f \in \czxr $ an input to this network. 
The action $\psi(f)$ can be consistely extended to an automorphism  ${\gamma}_{(\psi,f)} : S \rightarrow S$, \ie
${\gamma}_{(\psi,f)} \in \Gamma$.
\end{theorem}

The proof is given in the Appendix. A couple of notes. First, the 
input $f$ appears as a parameter for the automorphism (in addition to $\psi$),
as $\psi$ alone cannot define a consistent self-map over $S$. Second, 
this correspondence is not necessarily unique.
There's a family of automorphisms that can correspond to the action $\psi(f)$, but
we're interested in the {\em existence} of at least one of them.

\section{Group Actions underlying Deep Networks}
\label{sec:group}
\subsection{Shadow Stabilizer-subgroups}

We now search for group
actions that approximate the automorphisms we established.
Since such a group action is not exactly a neural network, yet can be closely mimics 
the latter, we will refer to these groups as {\em Shadow groups}.
The existence of an underlying group action asserts that corresponding to a set of stabilizers for a figure $f$ in
$\Omega$, there is a stabilizer subgroup, and lets us argue that the learnt figures
actually correspond to minimal orbits, with high probability - and thus the simplest possible.
The following theorem asserts this fact.

\begin{theorem}
\label{thm-grpnext}
 Let $\psi \in \Omega$ be a neural network working over a figure $f \subseteq S$, and the corresponding self-map
 $\gamma_{(\psi, f)} : S \rightarrow S$, then in fact $\gamma_{(\psi, f)} \in \mathrm{Homeo}(S)$, the homeomorphism group of $S$.
\end{theorem}

The above theorem (see Appendix, for a proof) shows that although neural networks may not exactly define a group,
they can be approximated well by a set of group actions - that of the homeomorphism group of the
intrinsic space. 
One can go further, and inspect the action of $\psi$ locally -
\ie in the small vicinity of each point. Our next result shows that locally, they can be approximated further by elements of $GL(S)$,
which is a much simpler group to study; in fact our results from section \ref{sec:gabor} were really 
in light of the action of this group for the 2 dimensional case. 

\subsection*{Local resemblance to $GL(S)$}
\begin{theorem}
\label{thm-grpdown}
For any $\gamma_{(\psi, f)} \in \homeo(S)$, there is a local approximation $g_{(\psi,f)} \in GL(S)$
that approximates $\gamma_{(\psi, f)}$. In particular, if $\gamma_{(\psi, f)}$ is a stabilizer for $f$, so is $g_{(\psi,f)}$.

\end{theorem}

The above theorem (proof in Appendix) guarantees an underlying group action.
But what if some large stabilizers in $\Omega$ are mapped to very small stabilizers in $GL(S)$, and vice versa?
The next theorem (proof in Appendix) asserts that there is a bottom-up correspondence as well - every group symmetry over the intrinsic space
has a counter-part over the set of mapping from $C_0(S, \real)$ onto itself. 

\begin{theorem}
\label{thm-omegaup}
Let $S$ be an intrinsic space, and $f \subseteq S$. Let $g_f \in GL(S)$ be a group element that 
stabilizes $f$. Then there is a map $ U : GL(S) \rightarrow \Omega$, such that the 
corresponding element $ U(g_f) = \tau_f \in \Omega$ that stabilizes
$f$. Moreover, for $g_1, g_2 \in GL(S)$, $U(g_1) = U(g_2) \Rightarrow g_1 = g_2$.
\end{theorem}

\textbf{Summary Argument :}
We showed via Theorem \ref{thm-grpnext} that any neural network action has a
counterpart in the group of homeomorphisms  over the intrinsic space.
The presence of this {\em shadow} group lets one carry over the 
orbit/stabilizer principle discussed in section \ref{sec:random-walk} to
the actual neural network transforms. Which asserts that the {\em simple}
features are the ones to be learned first. 
To analyse how these features look like, we can examine them
locally with the lens of $GL(S)$, an even simpler group.
The Theorems \ref{thm-grpdown} and \ref{thm-omegaup} collectively establish this.
Theorem \ref{thm-grpdown} shows that for every neural network element there's a nearby
group element. For a large stabilizer set then, the corresponding stabilizer subgroup
ought to be large, else it will be impossible to find a nearby group element everywhere.
Also note that this doesn't require a strict one-to-one correspondence; existence of
some group element is enough to assert this. 
To see how Theorem \ref{thm-omegaup} pushes the argument in reverse, imagine
a sufficiently discrete version of $GL(S)$. 
In this coarse-grained picture, any small $\epsilon$-neighbourhood of an element $g$
can be represented by $g$ itself (similar to how integrals are built up with small regions
taking on uniform values) . There is 
a corresponding neural network $U(g)$, and furthermore, for another element
$f$ which is outside this neighbourhood (and thus not equal to $g$) $U(f)$ is
another different network. This implies that a large volume
cannot be mapped to a small volume in general - and hence true for
stabilizer volumes as well.

\section{Deeper Layers and Moduli-space}
\label{sec:moduli}
Now we discuss how the orbit/stabilizer interplay extends to
multiple layers.  At its root, lies the principle of layer-wise pre-training for the identity function.
In particular, we show that this succinct step, as an algorithmic principle, is quite powerful. 
The principle stays the same across layers; hence every layer only learns the simplest possible objects from its own input space. Yet, a simple objects at a deeper layer can
represent a complex object at the input space.
To understand this precisely, we first examine the critical role of the sigmoid function.

\subsection{The role of the Sigmoid Function}
\label{sec:sigmoid}

Let's revisit the the transfer functions from section \ref{sec:convolution} (convolution view of a neuron):

\begin{minipage}{.5\linewidth}
\begin{equation}\tag{\ref{eqn:tau}}
   \tau_w( I) = \int \limits_{\theta \in X} w(\theta)I(\theta)d\theta
   \end{equation}
\end{minipage}
\begin{minipage}{.5\linewidth}
\begin{equation}\tag{\ref{eqn:zeta}}
    \zeta_w( I) = \sigma(\tau_w(I))
  \end{equation}
\end{minipage}

Let $\mathcal{F}_{\real}(A)$ denote the space of real functions on any space $A$.
Now, imagine a (hypothetical) space of infinite number of neurons indexed by
the weight-functions $w$. Note that $w$ can also be viewed as an  element of $\czxr$.
Plus $\tau_w(I) \in \real$. So the family $ \tau = \{ \tau_w\}$ induces a 
mapping from $\czxr$ to real functions over $\czxr$, \ie
\begin{equation}
\tau: \czxr \rightarrow \mathcal{F}_{\real}( \czxr)
\end{equation}
Now, the sigmoid can be thought of composed of two steps.
$\sigma_1$ turning its input to number between zero and one. And then, for most cases, an automatic thresholding
happens (due to discretized representation in computer systems) creating a binarization to 0 or 1. 
We denote this final step by $\sigma_2$, which reduces to the output to an element of the figure space $F(\czxr)$. 
These three steps, applying $\tau$, and then $\sigma = \sigma_1\sigma_2$ can be viewed as 
\begin{equation}
\label{eqn:point-to-figure}
\czxr \overset{\tau}{\rightarrow} \mathcal{F}_{\real}( \czxr)	\overset{\sigma_1}{\rightarrow}  \mathcal{F}_{[0,1]}( \czxr)   \overset{\sigma_2}{\rightarrow}     F(\czxr)
\end{equation}
This construction is recursive over layers, 
so one can envision a neural net building representations of moduli spaces over moduli spaces, hierarchically, one after another.
For example, at the end of layer-1 learning, each layer-1 neuron actually learns a figure over the
intrinsic space $\oxr$ (ref. equation \ref{eqn:tau}).  
And the collective output of this layer can be thought of as a figure over $\czxr$ (ref. equation \ref{eqn:point-to-figure}).
These second order figures then become inputs to layer-2 neurons, which collectively end up learning a minimal-orbit figure over the space of these second-order figures, and so on.

What does all this mean physically ? We now show that it is the power
to capture figures over moduli spaces, that gives us the ability to capture representation of features
of increasing complexity.

\subsection{Learning Higher Order Representations}
\begin{figure}
\hspace{-0.75in}\subfigure[Moduli spaces at subsequent layers]{\label {fig:multilayer}\includegraphics[width=3in, height=3in]{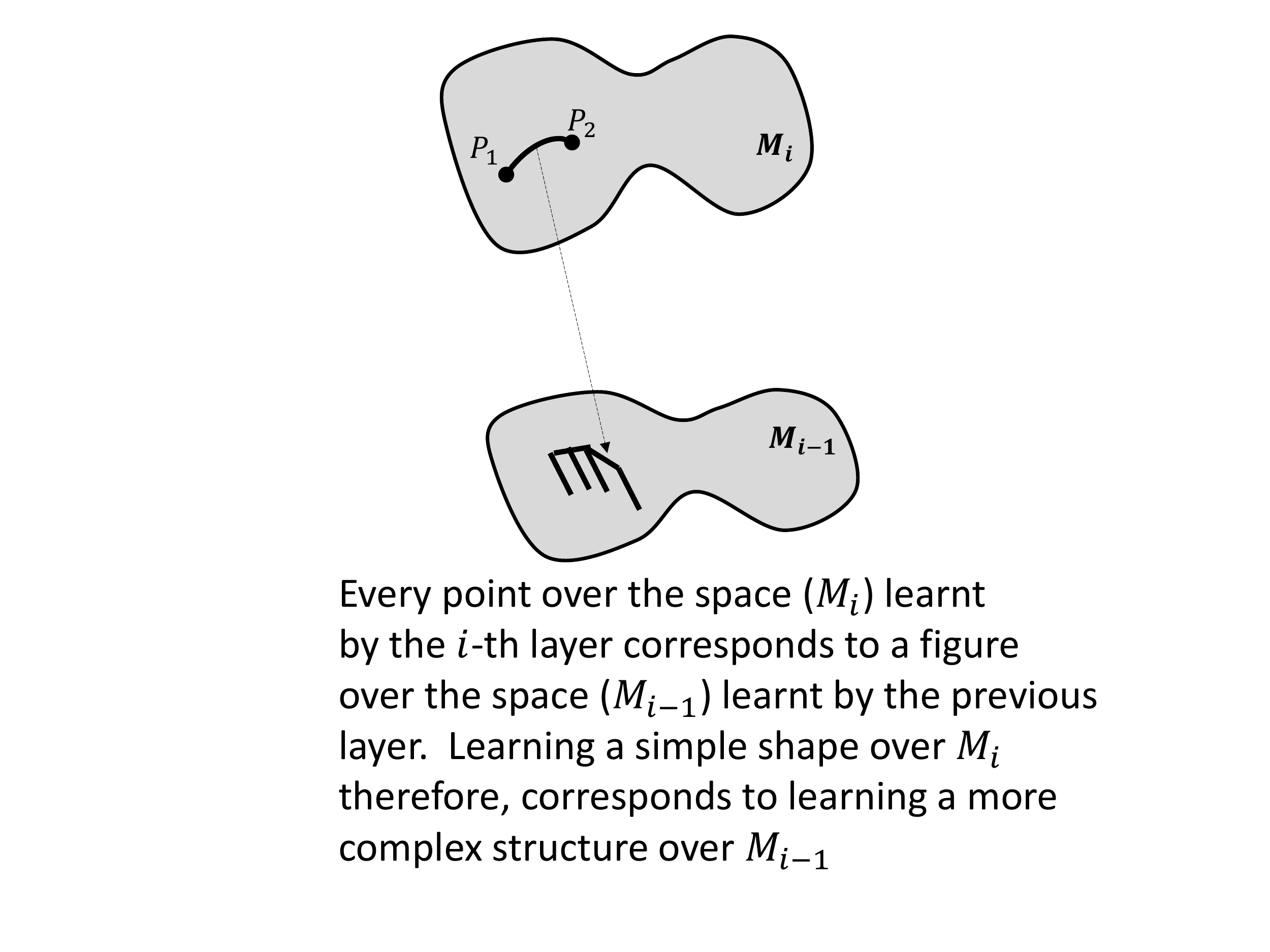}}
\hspace{-0.43in}\subfigure[Moduli space of line segments]{\label {fig:r4}\includegraphics[width=4in, height=3.5in]{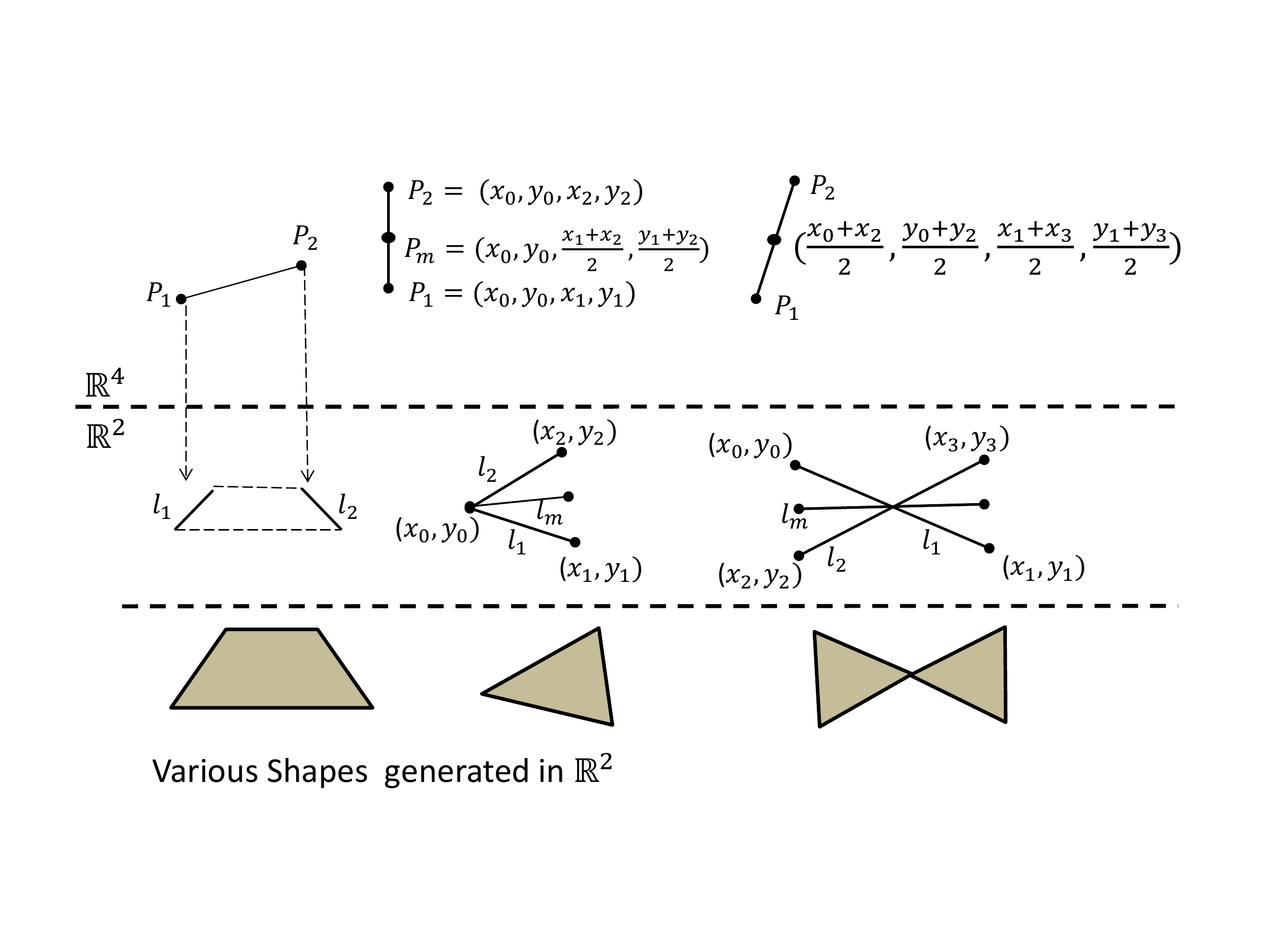}}
\caption{A deep network constructs higher order moduli spaces. It learns figures having increasingly smaller symmetry groups;
which corresponds to increasing complexity }
\end{figure}

Let's recap. A layer of neurons collectively learn a set of figures over its own intrinsic space. 
This is true for any depth, and the learnt figures correspond to the largest stabilizer subgroups. 
In other words, these figures enjoy highest possible symmetry. The set of simple examples in section \ref{sec:gabor} revealed that for $S = \plane$, the emerging figures are the edges. 

For a layer at an arbitrary depth, and working over a very different space, the corresponding figures
are clearly not physical edges. Nevertheless, we'll refer to them as \textbf{\em Generalized Edges}. 

Figure \ref{fig:multilayer} captures a generic multi-layer setting.
Let's consider the neuron-layer at the $i$-th level, and let's denote the embedding space of the
features that it learns as $M_i$ - i.e., this layer learns figures over the space $M_i$. 
One such figure, now refereed to as a {\em generalized edge}, is schematically shown by the {\em geodesic segment} $P_1P_2$ in the 
picture. Thinking down recursively, this space $M_i$ is clearly a moduli-space of {\em figures} over
the space $M_{i-1}$, so each point in $M_i$ corresponds to a generalized edge over $M_{i-1}$.
The whole segment $P_1P_2$ therefore corresponds to a collection of generalized edges
over $M_{i-1}$; such a collection in general can clearly have lesser symmetry than a 
generalized edge itself. In other words, a simple object $P_1P_2$ over $M_i$ actually corresponds to
a much more complex object over $M_{i-1}$.

These moduli-spaces are determined by the underlying input-space, and the nature of training.
So, doing precise calculations over them, such as defining the space of all automorphisms, 
or computing volumes over corresponding stabilizer sets, may be very difficult, and we 
are unaware of any work in this context. However, the following simple example
illustrates the idea quite clearly.

\subsubsection{Examples of Higher Order Representation}
Consider again the intrinsic space $\plane$. An edge on this plane is a line segment - $[(x_1,y_1), (x_2, y_2)]$.
The moduli-space of all such edges therefore is the entire 4-dimensional real Euclidean space sans the origin - $\real^4/\{0\}$. Figure \ref{fig:r4} captures this. Each point in this space ($\real^4$) corresponds to an edge over the plane ($\plane$). A generalized-edge over $\real^4/\{0\}$, which
is a standard line-segment in a 4-dimensional euclidean space, then
corresponds to a collection of edges over the real plane. Depending on the orientation
of this {\em generalized edge} upstairs, one can obtain many different shapes downstairs.

\textbf{The Trapezoid }The figure in the leftmost column of figure \ref{fig:r4} shows 
a schematic trapezoid. This is obtained from two points $P_1, P_2 \in \real^4$ that
correspond to two non-intersecting line-segments in $\real^2$. 

\textbf{A Triangle} The middle column shows an example where the starting point of the two
line segments are the same - the connecting {\em generalized edge between} the two points in $\real^4$
is a line parallel to two coordinate axes (and perpendicular to the other two). 
A middle point $P_m$ maps to a line-segment that's intermediate between 
the starting lines (see the coordinate values in the figure). Stabilizing $P_1P_2$ upstairs effectively
stabilizes the the triangular area over the plane. 

\textbf{A Butterfly} The third column shows yet another pattern, drawn by two edges that intersect with each 
other. Plotting the intermediate points of the generalized edge over the plane as lines, a 
butterfly-like area is swept out. Again, stabilizing the {\em generalized edge} $P_1P_2$ effectively
stabilizes the butterfly over the plane, that would otherwise take a long time to be found and stabilized 
by the first layer.

\textbf{More general constructions} We can also imagine even more generic shapes that can be learnt this way.
Consider a polygon in $\real^2$. This can be thought of as a collection of triangles (via triangulation), and
each composing triangle would correspond to a {\em generalized edge} in layer-2. As a result, the whole 
polygon can be effectively learnt by a network with two hidden layers.

In a nutshell - the mechanism of finding out maximal stabilizers over the moduli-space of figures works uniformly
across layers. At each layer, the figures with the largest stabilizers are learnt as the candidate features.
These figures correspond to more complex shapes over the space learnt by the earlier layers. By adding deeper
layers, it then becomes possible to make a network learn increasingly complex objects over its originial input-space.

\section{Related Work}
\label{sec:related-work}
\vspace{-0.5cm}

This starting influence for this paper were the key steps described
by \cite{hinton2006reducing}, where the
authors first introduced the idea of layer-by-layer pre-training through
autoencoders. The same principles, but over Restricted Boltzmann machines (RBM),
were applied for image recognition in a later work (see \cite{salakhutdinov2009deep}).
\cite{lee2009convolutional} showed, perhaps for the first time,
how a deep network builds up increasingly complex representations
across its depth. Since then, several variants of autoencoders, as well as RBMs have taken the center stage of the deep learning research. \cite{bengio--dlbook} and  \cite{bengio-corr-1305-0445}
provide a comprehensive coverage on almost every aspect of DL techniques. 
Although we chose to analyse auto-encoders in this paper, we believe that the same 
principle should extend to RBMs as well, especially in the context
of a recent work by \cite{kamyshanska2014}, that reveals {\em seemingly} an equivalence between autoencoders and RBMs. 
They define an energy function
for autoencoders that corresponds to the free energy of an RBM. They also point out
how the energy function imposes a regularization in the space.
The hypothesis about an implicit regularization
mechanism was also made earlier by \cite{erhan2010does}. Although we haven't investigated any direct connection between 
symmetry-stabilization and regularization, there are evidences that they may
be connected in subtle ways (for example, see \cite{shah2012group}). 

Recently \cite{AnselmiLRMTP13} proposed a theory for visual-cortex that's heavily inspired 
by the principles of group actions; although there is do direct connection with layer-wise
pre-training in that context. \cite{BouvrieNIPS2009_3732}
studied invariance properties of layered network in a group theoretic framework and showed how to derive
precise conditions that must be met in order to achieve invariance - this is very close to
our work in terms of the machineries used, but not about how a unsupervised learning
algorithm learns representations.
\cite{mehta2014exact} recently showed an intriguing connection
between Renormalization group flow \footnote{This subject is widely studied in many areas of physics, such as quantum field theory, statistical mechanics and so on} and deep-learning. They constructed an explicit mapping from a renormalization group over a block-spin Ising model (as proposed by \cite{kadanoff1976variational}), to a DL architecture.
On the face of it, this result is complementary to ours, albeit in a
slightly different settings. Renormalization is a process of coarse-graining a system by
first throwing away small details from its model, and then examining the new system under the simplified model (see \cite{cardy1996scaling}). In that sense the orbit-stabilizer principle is a re-normalizable theory - it allows for the exact same coarse-graining operation at every layer - namely, keeping only minimal orbit shapes and then passing them as new parameters for the next layer - and the theory remains unchanged at every scale.

While generally good at many recognition tasks, DL networks 
have been shown to fail in surprising ways. \cite{szegedy2013intriguing}
showed that the mapping that a DL network learns could have 
sudden discontinuities. For example, sometimes it can misclassify an
image that is derived by applying only a tiny perturbation to
an image that it actually classifies correctly.
Even the reverse was also reported (see \cite{nguyenFoolDL2014}) - here, a
DL network was tested on grossly perturbed versions of already learnt images - perturbed to the extent that humans cannot recognize them for the original any more - and they were still classified as
their originals. \cite{szegedy2013intriguing} made a related observation:
random linear combination of high-level units in a deep network also
serve as good representations. They concluded - it is the space, rather than the individual
units, that contain the semantic information in the high layers of a DL network.
We don't see any specific conflicts of any of these observations
with the orbit-stabilizer principle, and view the possible explanations of these
phenomena in the clear scope a future work.

\vspace{-0.5cm}
\section{Conclusions and Future Work}
\vspace{-0.5cm}
In a nutshell, this paper builds a theoretical framework for unsupervised DL that is primarily inspired by
the key principle of finding a generative model of the input samples first. The framework is based on orbit-stabilizer interplay in group actions. We assumed {\em layer-wise} pre-training, since it
is conceptually clean, yet, in theory, even if many layers were being learnt simultaneously
(in an unsupervised way, and still based on reconstructing the input signal)
the orbit-stabilizer phenomena should still apply. We also analysed how higher order representations emerge as the networks get deeper. 

Today, DL expanded well beyond the principles \pone and \ptwo  (ref. introduction). 
Several factors such as the size of the datasets, increased computational power,
improved optimization methods, domain specific tuning, all contribute to its success.
Clearly, this theory is not all-encompassing. 
In particular, when large enough labeled datasets are available, training them in
fullly supervised mode yielded great results. The orbit-stabilizer principle cannot be
{\em readily extended} to a supervised case; in the absence of a self-map (input reconstruction)
it is hard to establish a underlying group action. But we believe and hope that 
a principled study of how representations form will eventually put the two under a single theory.

\section{Appendix}

This section involves elementary concepts from different areas of mathematics. 
For functional analysis see \cite{bollobás1999linear}. For elementary ideas on groups, Lie groups, and
representation theory, we recommend \cite{artin1991algebra, fulton1991representation}. The relevant ideas in topology
can be looked up in \cite{munkres1974topology} or \cite{hatcher2002algebraic}.

\begin{figure}
\includegraphics[width=5in, height=3in]{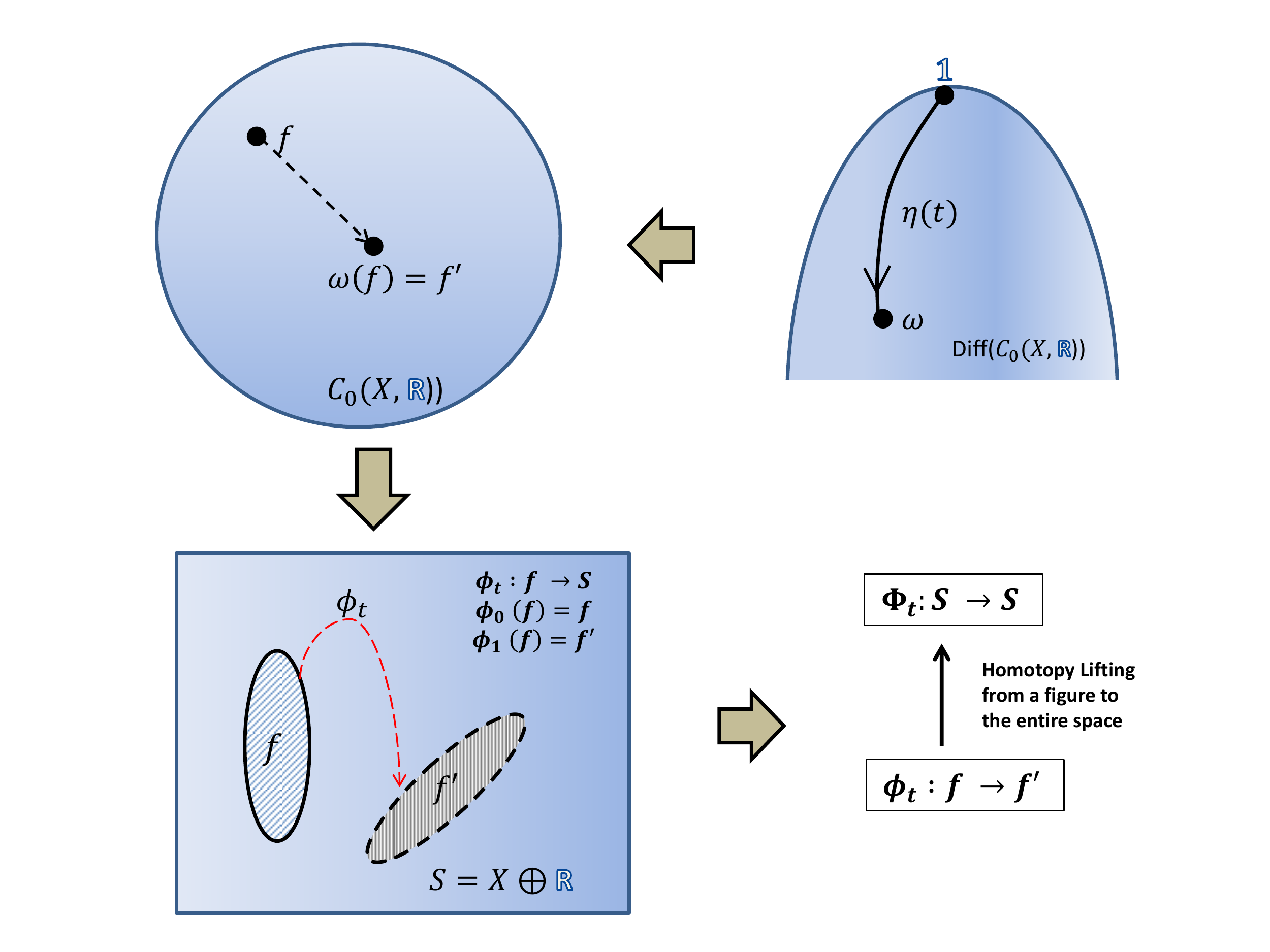}
\caption{ Homotopy Extension over the intrinsic space $S$. An imaginary curve (geodesic) $\eta$ over the diffeomorphisms
induces a family of continuous maps from $f$ to $f^\prime$ in $S$. This homotopy can be extended to the 
entire space $S$ by the homotopy extension theorem.  }
\label{fig:lifting}
\end{figure}

\begin{lemma}
\label{lem-dense}
  The group of invertible square matrices are dense in the set of square matrices.
\end{lemma}

\textbf{Proof} -  This is a well known result, but we provide a proof for the sake of completeness.
Let $A$ be $n \times n$ matrix, that is square, and not necessarily invertible. We show that 
there is a non-singular matrix nearby. To see this, consider an arbitrary non-singular matrix $B$, \ie $\text{det}(B) \neq 0$,
and consider the following polynomial parametrized by a real number $t$,
\[
	r(t) = \text{det} ((1-t)A + tB))
\]

Since $r$ is finite degree polynomial, and it is certainly not
identically zero, as $r(1) = det(B) \neq 0$, it can only vanish at a finite number of points. So,
even if $p(0) = 0$, there must be a $t'$ arbitrarily close to $0$ such that $r(t') \neq 0$.
So the corresponding new matrix $M = (1-t')A + t'B $ is arbitrarily close to $A$, yet non-singular, as $\text{det}(M) = r(t')$ does not identically vanish. $\Box$

\begin{lemma}
\label{lem-invertible}
  The action of a neural network  can be approximated by a network that's completely invertible.
\end{lemma}

\textbf{Proof} - A three layer neural network can be represented as $W_2 \sigma W_1$, where 
$\sigma$ is the sigmoid function, and $W_1$ and $W_2$ are
linear transforms. $\sigma$ is already invertible, so we only need to show existence
of invertible approximations of the linear components. 

Let $W_1$ be a $\real^{m \times n}$ matrix. Then $W_2$ is a matrix of dimension $\real^{n \times m}$. 
Consider first the case where $m > n$. The map $W_1$ however can be lifted to a $\real^{m \times m}$ transform, with
additional $(m-n)$ columns set to zeros. Let's call this map $W_1^\prime$.  But then, by 
Lemma \ref{lem-dense}, it is possible to obtain a square invertible transform $\overline{W_1}$ 
that serves as a good approximation for $W_1^\prime$, and hence for $W_1$. One can obtain a similar
invertible approximation $\overline{W_2}$  of $W_2$ in the same way, and thus the composite map $\overline{W_2}\sigma \overline{W_1}$   
is the required approximation that's completely invertible. 
The other case where $m < n$ is easier; it follows from the same logic
without the need for adding columns. $\Box$

\begin{lemma}
\label{lem-setmapping}
  Let the action of a neural network $\psi$ over a figure $f$ be $\psi(f) = f^\prime$ where $f, f^\prime \subseteq S$, the intrinsic space.
  The action then induces a continuous mapping of the sets $f \xrightarrow{\psi} f^\prime$.  
\end{lemma}

\textbf{Proof} 
Let's consider an infinitely small subset $\epsilon_a \subset f$. Let $f_a = f \backslash  \epsilon_a$.
The map $\psi(f)$ can be described as
$\psi(f) = \psi ( \{f_a \cup \epsilon_a\}) = f^\prime$. Now let's imagine an infinitesimal
deformation of the figure $f$ - by continuously deforming $\epsilon_a$ to a new set $\epsilon_b$.
Since the change is infinitesimal, and $\psi$ is continuous, the new output, let's call it  $f^{\prime\prime}$, differs from $f^\prime$ only infinitesimally. That means there's a subset $\epsilon_a^\prime \subset f^\prime$
that got deformed into a new subset $\epsilon_b^\prime \subset f^{\prime\prime}$ to give rise to $f^{\prime\prime}$.
This must be true as Lemma \ref{lem-invertible} allows us to assume an invertible mapping $\psi$ - so that
a change in the input is  detectable at the output. Thus $\psi$ induces a mapping between the infinitesimal sets 
$\epsilon_a \xrightarrow{\psi} \epsilon_a^\prime$. Now, we can split an input figure into infinitesimal
parts and thereby obtain a set-correspondence between the input and output. $\Box$

\subsubsection*{ Proof of Theorem \ref{thm-automorph}}
We assume that a neural network implements a differentiable map $\psi$ between its input and output.
That means the set $\Omega \subseteq \mathrm{Diff}(\czxr)$ - \ie the set of diffeomorphisms of $\czxr$.
This set admits a smooth structure (see \cite{ajayi97}). Moreover, being parametrized by the neural-network parameters, the set is connected
as well, so it makes sense to think of a curve over this space.
Let's denote the identity transform in this space by $\id$. 
Consider a curve $\eta(t) : [0,1] \rightarrow \diffczxr$,
such that $\eta(0) = \id$ and $\eta(1) = \psi$. By Lemma \ref{lem-setmapping}, 
$\eta(t)$ defines a continuous map between $f$ and $\eta(t)(f)$.
In other words, $\eta(t)$ induces a partial homotopy on $S = \oxr$.
Mathematically, ${\eta(0)} =\id(f) = f$ and ${\eta(1)} =\psi(f) = f^\prime$.
In other words, the curve $\eta$ induces a continuous family of deformations of $f$ to $f^\prime$. 
Refer to figure \ref{fig:lifting} for a visual representation of this correspondence.
In the language of topology such a family of deformations is known as a homotopy.
Let us denote this homotopy by $\phi_t : f \rightarrow f^\prime$.
Now, it is easy to check (and we state without proof) that for a $f \in \czxr$,
the pair $(\oxr, f)$ satisfies Homotopy Extension property.  In addition, there
exists an initial mapping $\Phi_0 : S \rightarrow S$ such that $\Phi_0|_f = \phi_0$.
This is nothing but the identity mapping. The well known Homotopy Extension Theorem (see \cite{hatcher2002algebraic}) then
asserts that it is possible to lift the same homotopy defined by $\eta$ from $f$ to the entire set $\oxr$. In other words, 
there exists a map $\Phi_t : \oxr \rightarrow \oxr$, such that the restriction $\Phi_t|_f = \phi_t$ for all $t$.
Clearly then,  $\Phi_{t=1}$  is our intended automorphism on $S = \oxr$, that
agrees with the action $\psi(f)$. We denote this automorphism by $\gamma_(\psi, f)$, and this
is an element of $\Gamma$, the set of all automorphisms of $S$.
 $\Box$ \\

\subsubsection*{ Proof of Theorem \ref{thm-grpnext}}
Note that lemma \ref{lem-invertible} already guarantees invertible mapping
beteen $f$ and $f^\prime = \psi(f)$. Which means in the context of
theorem \ref{thm-automorph}, there is an inverse homotopy, that can be
extended in the opposite direction. This means $\gamma_{(\psi, f)}$
is actually a homeomorphism, \ie a continuous invertible mapping from $S$ to itself. 
The set $\mathrm{Homeo}(S)$ is a group by definition.$\Box$

\subsubsection*{ Proof of Theorem \ref{thm-grpdown}}
Let $\psi(f)$ be the neural network action under consideration.
By Theorem \ref{thm-automorph},  there is a corresponding homeomorphism $\gamma_{(\psi, f)} \in \Gamma(S)$.
This is an operator acting $S \rightarrow S$, and although not necessarily differentiable everywhere,
the operator (by its construction) is  differentiable on $f$. This means it can
be locally (in small vicinity of points near $f$) approximated by its Fr\'echet derivative 
(analogue of derivatives over Banach space ) near $f$; however, in finite dimension
this is nothing but the Jacobian of the transformation, which can be represented by 
a finite dimensional matrix. So we have a linear approximation of this deformation $\gamma_{(\psi, f)}   = J (\gamma_{(\psi, f)} ) = \hat{\gamma}$.
But then since this is a homeomorphism, by the inverse function theorem, 
$\hat{\gamma}^{-1}$ exists. Therefore $\hat{\gamma}$ really represents an element 
$g_\psi \in GL(S)$. $\Box$

\subsubsection*{ Proof of Theorem \ref{thm-omegaup}}
 Let $S$ be the intrinsic space over an input vector space $X$, \ie $ S = X \times \real$,
 and $g_f \in GL(S)$ a stabilizer of a figure $f \in \czxr $. 
 Define a function $\chi_f : F(S) \rightarrow F(S)$
 \[
 \chi_f(h) = \bigcup\limits_{s \in h} g_f(s) = h^\prime
 \]
 
 It is easy to see that $\chi_f$ so defined stabilizes $f$.  However there is the possibility that 
- $h^\prime \notin \czxr$; although $h^\prime$ is a figure in $S$, it may not be a well-defined function over $X$. To avoid such a pathological case, we make one more 
assumption - that the set of functions under considerations are all bounded.
That means - there exists an upper bound $B$, such that every $f \in \czxr$ under consideration is bounded in supremum norm - $|f|_{supp} < B$ \footnote{This is not a restrictive assumption, in fact
it is quite common to assume that all signals have finite energy, which implies that the signals
are bounded in supremum norms}.
Define an auxiliary function $\hat{\zeta}_f : X \rightarrow \real$ as follows.
\begin{align*}
	\hat{\zeta_f} (x) = B  \;\;\;\; \forall x \in support(f) \\
	 \hat{\zeta}_f (x) = 0  \;\;\;\;\text{otherwise}
\end{align*}

Now, one can always construct a continuous approximation of $\hat{\zeta_f}$ - let $\zeta_f \in \czxr$ be such an approximation. We are now ready to define the neural network $U(g_f) = \tau_f$. Essentially it is the collection
of mappings  between figures (in $\czxr$) defined as follows: 
\begin{align*}
	\tau_f(h) = \chi_f(h)	 \;\; \text{whenever} \;\;\; \chi_f(h) \in \czxr \\
	\tau_f(h) = \zeta_f \;\;\;\;\text{otherwise} \hspace{70pt}
\end{align*}

To see why the second part of the theorem holds, observe that
since $U(g_1)$ and $U(g_2)$ essentially reflect group actions over the
intrinsic space, their action is really defined point-wise. 
In other words, if $p,q \subseteq S$ are figures, and $r = p \cap q$,
then the following restriction map holds.

\[
	U(g_1)(p)|_r = U(g_1)(q)|_r
\]

Now, further observe that given a group element $g_1$, and a point $x \in S$, one can always construct a family of figures  containing $x$ all of which are valid functions in $\czxr$, and 
that under the action of $g_1$ they remain valid functions in $\czxr$. Let $f_1,f_2$
be two such figures such that $f_1 \cap f_2 = x$. 
Now consider $x^\prime = U(g_1)(f_1) \cap
U(g_1)(f_2)$.  However, the collection of mapping $\{ x \rightarrow x^\prime \}$ uniquely
defines the action of $g_1$. So, if $g_2$ is another group element
for which $U(g_2)$ agrees with $U(g_1)$ on every figure, that agreement can
be translated to point-wise equality over $S$, asserting $g_1 = g_2$. $\Box$

\subsection{ A Note on the Thresholding Function}
In section \ref{sec:sigmoid} we primarily based our discussion around the sigmoid function that can
be thought of as a mechanism for binarization, and thereby produces figures
over a moduli space. This moduli space then becomes an intrinsic space for the next layer.
However, the theory extends to other types of thresholding functions as well, only the intrinsic
spaces would vary based on the nature of thresholding.
For example, using a linear rectification unit, one would get a mapping into the space $\mathcal{F}_{[0,\infty]}(\czxr)$.
The elements of this set are functions over $\czxr$ taking values in the range $[0,\infty]$.
So, the new intrinsic space for the next level will then be $S = [0,\infty] \times \czxr]$, and the output
can be thought of as a figure in this intrinsic space allowing the rest of the theory to carry over.


\bibliography{geom_dl_paper}
\bibliographystyle{iclr2015}

\end{document}